\pdfoutput=1
\documentclass[twocolumn]{svjour3}          
\usepackage{graphicx}
\usepackage{amssymb}
\usepackage{amsmath}
\usepackage{amsfonts}
\usepackage{float}
\usepackage{subcaption}
\usepackage[justification=centering]{caption}
\usepackage{graphicx}
\usepackage[dvipsnames]{xcolor}
\usepackage{changes}
\usepackage[numbers]{natbib}

\journalname{Signal, Image and Video Processing}
\begin{document}

\title{Just Noticeable Difference for Machine
Perception and Generation of Regularized Adversarial Images with Minimal
Perturbation
}


\author{Adil Kaan Akan       \and
        Emre Akbas \and
        Fatos T. Yarman Vural 
}


\institute{$1\quad$ Computer Engineering Department, Middle East Technical University, Ankara, Turkey \\
              \email{\{kaan.akan, emre, vural\}@ceng.metu.edu.tr}           %
}

\date{}

\titlerunning{JND for Machine
Perception and Generation of Regularized Adversarial Images with Minimal
Perturbation}
\maketitle


\begin{abstract}
In this study, we introduce a  measure for machine perception, inspired by the concept of Just Noticeable Difference (JND)  of human perception. Based on this measure, we suggest an adversarial image generation algorithm,  which iteratively distorts an image by an additive noise until the model detects the change in the image by outputting a false label. The noise added to the original image is defined as the gradient of the cost function of the model. A novel cost function is defined to explicitly minimize the amount of perturbation applied to the input image while enforcing the perceptual similarity between the adversarial and input images. For this purpose, the cost function is regularized by the well-known total variation and bounded range terms to meet the natural appearance of the adversarial image.  We evaluate the adversarial images generated by our algorithm both qualitatively and quantitatively on  CIFAR10, ImageNet, and MS COCO datasets. Our experiments on image classification and object detection tasks show that adversarial images generated by our JND method are both more successful in deceiving the recognition/detection models and less perturbed compared to the images generated by the state-of-the-art methods, namely, FGV, FSGM, and DeepFool methods.
\end{abstract}

\section{Introduction}
 Contemporary Computer Vision algorithms model the visual information by deep neural networks, which yield human-like performances in some applications, such as face recognition and image classification \cite{6909616,he2015delving}. However, these algorithms fail unexpectedly for intentionally designed test images even if they look very similar to the images in the training data set \cite{goodfellow2014explaining, rozsa_adversarial_2016, moosavi-dezfooli_deepfool_2016}. Such examples are called adversarial images, and they create vital problems in high-technology products.  Unfortunately, most of the current image classifiers and/or object detectors are not secure enough against these types of deceptions. One way to support the efforts towards robustness is to design effective algorithms that generate adversarial images with lower perturbations and a higher success rate in deceiving the models.

Applying a large perturbation to an image, e.g., by completely changing it with another image, would not be considered as a valid adversarial example generation method since the notion of minimal perturbation is implied in the definition of the adversarial image. For example, the Fast Gradient Sign Method (FGSM) \cite{goodfellow2014explaining} inherently uses the $L_\infty$ norm by searching for a minimal $\epsilon$ (magnitude of change), and the DeepFool \cite{moosavi-dezfooli_deepfool_2016} method explicitly minimizes the $L_2$ norm of the perturbation. However, keeping the amount of perturbation minimal to preserve the perceptual similarity to the original image as much as possible has not been the main focus of these state-of-the-art studies. In particular,  regularization techniques have not been explored to minimize perceptual perturbation.

In this paper, we propose a method called, Just Noticeable Difference (JND)  to generate adversarial images as an extension and generalization of our preliminary work \cite{9191090}. The JND method is a gradient-based white-box adversarial attack that focuses on preserving the perceptual similarity to the original image by using image regularization techniques while minimizing the perturbation. Inspired by experimental psychology, we adopt the concept of Just Noticeable Difference of human perception to machine perception. We define JND for a  machine learning model as the just noticeable distortion of an image to cheat a machine learning algorithm. In order to keep the distortion as small as possible and consistent with human perception, we define a perceptually meaningful additive noise, which is proportional to the gradient of the cost function of the model. The cost function is enriched by two regularization terms, namely, bounded range and total variation  \cite{mahendran_visualizing_2016} to assure that the adversarial images are as close as possible to the original images with respect to human perception. In other words, the regularization terms enforce the machine perception to human perception to a certain extent. 

The major contribution of this study is to introduce the concept of  Just Noticeable Difference, $\mathrm{JND}(M)$ for a machine learning model, $M$, which is associated with a novel cost function regularized by additional terms to reduce the discrepancy between machine and human perception. A formal definition of Just Noticeable Difference is provided for machine perception. A true image is distorted by the gradient descent method iteratively with a small additive noise proportional to the cost function's gradient to capture the JND adversarial image.

This manuscript involves a substantial extension and generalization of our previous work \cite{9191090}, where we formalized the concept of Just Noticeable Difference for the generation of adversarial images. Our extension includes a mathematical definition of JND($M$) for a machine learning model, $M$. We also, conduct thorough  experiments to analyze the power of the suggested method and comparisons to the state of the art adversarial image generation techniques,  as summarized below: 

\begin{itemize}
   
    \item We  compare our method  with the well-known adversarial image generation algorithms, namely,  FGSM \cite{goodfellow2014explaining}, FGV \cite{rozsa_adversarial_2016} and DeepFool \cite{moosavi-dezfooli_deepfool_2016} methods, in terms of (i) generation speed, (ii) image quality and  (iii) distance to the original image.
    \item We analyze the statistical similarities of the original and adversarial images by estimating the distribution of the Kullback-Leibler divergence and that of the $L_2$ distances between the adversarial and original images. 
    
    \item We introduced a non-targeted version of our method and compared its performance to that of FGSM and FGV methods. 
    
\end{itemize}

\section{Major Approaches for Adversarial Image Generation}
\label{sec:rw}

The vast amount of adversarial image generation methods reported in the literature can be grouped under two headings;  i) approaches that modify the whole image and ii) approaches that modify only a few pixels.  Adversarial images generated by modifying the whole image are mainly used for augmenting the training sets. On the other hand,  adversarial images generated by modifying only a few pixels do not contribute to improve the statistical properties of the dataset for improving the model. However, they have crucial merit to show the vulnerability of the models to spurious images.

The type of adversarial attacks can be grouped under two headings; i) targeted attacks and ii) non-targeted attacks. In targeted attacks, the method aims to trick the model into a specific target label. In order to achieve this aim,  the image is gradually changed so that the model recognizes the specific target label. However, in non-targeted attacks, the method aims to fool the model without the specific target label. This aim is achieved by gradually changing the image until the model outputs a label different than the true one.

 In this study, we suffice to provide a few pioneering methods for adversarial image generation and attacks, as summarized below.

\subsection{Approaches that modify the whole image}

Adversarial images generated by modifying the whole image are more likely to contribute to the estimation of the class distributions during the training phase of a machine learning model. Therefore, images generated under this approach are more suitable to improve the robustness of the models compared to the few pixel modification approaches, summarized in the next subsection. Since the suggested JND method modifies the whole image, we suffice to experimentally compare it to the methods in this group.

In most cases, humans cannot recognize the changes in an adversarial example, but a model may easily discriminate it from the true image and may assign it to another category. Kurakin et al. \cite{kurakin_adversarial_2016} state that in the physical world scenarios, adversarial examples are a crucial threat to the classifiers. They observe this problem by using adversarial examples, which are obtained from a cell phone camera. They feed these images to an ImageNet pre-trained Inception classifier and measure the accuracy of the classifier. They observe that most of the adversarial examples are misclassified.

A study by Szegedy et al. found some intriguing properties of neural networks \cite{szegedy_intriguing_2013}. For example, the adversarial examples generated on ImageNet were very similar to the original ones, so that even the human eye failed to distinguish them. Interestingly, some of the images also got misclassified by other classifiers that had different architectures, or they were trained on different subsets of the training data. These findings sadly suggest that deep learning classifiers, even the ones that obtain superior performances on the test set, do not actually learn the true underlying patterns that determine the correct output label. Instead, these algorithms built a model that works well on frequently occurring data but fails miserably for the data that do not have a high probability in the distribution. Szegedy et al. also proposed a box constrained attack, which tries to change the label of a true image by adding noise using the L-BFGS method.

Papernot et al. claimed that the adversarial attacks require the knowledge of either the model internals or its training data
\cite{papernot_practical_2017}. They introduce a practical method of an attacker, controlling a remotely hosted deep neural network. The only capability of the black-box adversary generator is to observe labels given by the deep neural network to selected inputs. 

Moosavi-Dezfooli et al. proposed a method called  DeepFool \cite{moosavi-dezfooli_deepfool_2016}, which aims to find the "closest" adversarial example to the original image with respect to $L_p$ norms. The model generates a perturbation for all classes and selects the one, which makes the adversarial image the closest adversarial example to the original image. They also proposed a universal adversarial perturbations method \cite{moosavi2017universal}, which estimates an adversarial perturbation from multiple images to fool the model with any image.  The goal of the algorithm is to find a universal adversarial perturbation, such that when it is added to an image, it fools the model. 

Goodfellow et al. proposed an adversarial image generation method \cite{goodfellow2014explaining}, called  Fast Gradient Sign Method (FGSM), which performs a one-step update on the input image using the sign of the calculated gradient.  However, there is a trade-off in the method, which requires a well-adjusted epsilon value to determine the step size. If the epsilon value is too low, the attack success rate is not sufficient, and if the epsilon value is too high, the naturalness of the images suffers from it. 

Rozsa et al. proposed the Fast Gradient Value (FGV) method \cite{rozsa_adversarial_2016}, as a modification to FGSM. In FGV, instead of using the sign of the gradient, they used its magnitude.  
Moreover, they proposed the "hot/cold approach" as a targeted adversarial attack, which takes advantage of the feature map of the Convolutional Neural Network. In the hot/cold approach, authors chose two classes, one is the hot class, which they want the model to approach, and the other is the cold class which they want the model to move away. Then, they minimize the distance between convolutional feature maps of the hot class and maximize the distance between convolutional feature maps of the cold class.

Note that none of the above-mentioned methods enforce the perceptual similarity of the adversarial and the real images. Furthermore, most of the methods do not explicitly minimize the amount of perturbation of the adversarial images in their objective function. The only exception is the DeepFool method. However, this method is not cautious about the naturalness of the generated images.

\subsection{Approaches that modify only a few pixels}

The methods under this group generate adversarial images that differ from the original image with a minimum number of pixels. Although the methods are very critical in showing the vulnerability of the models to slight modifications, the generated adversarial images may lack statistically significant contributions to improve the estimation of the distributions. In the following, we briefly summarize this group of approaches for completeness.

Su et al. and Vargas et al. introduced "one pixel attacks"  \cite{su2019one} \cite{vargas2019understanding}. Instead of changing the whole image, these methods proposed to change only a single pixel in the image. They successfully trick the attacked models. However,  in most cases, the change in one pixel looks like an artifact and is detectable by the human eye.

Papernot et al. proposed changing only a few pixels of the image instead of changing the whole \cite{papernot2016limitations}. Their algorithm changes one pixel of the image at a time and monitors the results of the change by checking the results of the classification. The algorithm computes a saliency map using the gradients of the outputs of the network layers. Based on this saliency map, the algorithm chooses the most effective pixel by choosing the largest value in the saliency map, which indicates a higher probability of fooling the network.

\section{Just Noticeable Difference for Machine Perception}

In experimental psychology, Just Noticeable Difference (JND) is defined as the least amount of variation on some sensory stimuli in order to detect the change. In this study, we adopt this concept to machine perception, where the machine detects the change on the input image and decides that it is different from the original image. 

Let us start by defining the concept of Just Noticeable Difference (JND) for a Machine Learning model. This is a  critical step to generate an adversarial image from a real image,  which confounds a machine learning algorithm. The adversarial image with the \textbf{least perceptible difference}, in which the network discriminates the true image from the adversarial image is called \textit{just noticeably different (JND)  adversarial image}. The formal definition of JND is given below:

\textbf{Definition: Just Noticeable Difference for a Machine Learning Model:} Suppose that a machine learning model $M$, generates a true label $y$ for  image $x$,
 $$M(x) = y.$$ 
 
 Suppose, also, that image $x$ is distorted gradually by adding an incremental noise $n(k)$ to generate an image 
 $$x(k+1)= x(k)+n(k),$$ at iteration $k$, where $x=x(0)$.

 Assure that $M[x(k)] = y,$ for all $k= 1,...., (K-1)$ and $M[x(k)] \neq  y $, for all  $k\ge K.$   

Given an image $x$, JND for a machine learning model $M$ is, then, defined as the difference image,
 \begin{equation}
 \label{eq:def_jnd}
  \mathrm{JND} [M(x)] =| x(0)-x(K)| ,
\end{equation}

\noindent where $|\cdot|$ indicates an element wise similarity metric between the true image $x(0)$ and the adversarial image $x(K)$. In the above definition, the model, $M$ detects the  perceptual change in the generated image $x(K)$ for the first time, at iteration $K$  and outputs a false label. 

\textbf{Definition: Just Noticeably Different Adversarial Image} is defined as the  image $x(K)$, where the machine learning model $M$ outputs a false class label for the first time at iteration $K$:

$$M[x(K)] \neq  y.$$

Note that $K$ is the\textbf{ smallest number of iteration step}, where the machine notices the difference between the original image and the adversarial image, as we gradually distort the original image. Note, also, that the image generated at iteration K, is the least detectable difference from the true image $x$ by the model $M$, because; for all the generated images, $x(k)$ for $k\le (K-1)$,   $M[x(k)] = y$. The rest of the images $x(k)$  for  k $ \ge K $  are adversarial, i.e.  $M[x(k)] \neq  y $. 

The crucial question is how  to generate  JND  adversarial images, which trick the model, yet perceptually just noticeably different from the  original image.

\section{Generating the Adversarial image by Just Noticeable Difference}

In order to generate an adversarial image with Just Noticeable Difference, $\mathrm{JND}(M)$, for a machine learning model $M$, we adopt the Convolutional Neural Network (CNN) architecture, suggested by  Gatys et al. \cite{gatys_image_2016} for style transfer. In their method, the final style image is obtained by starting from a random noise image fed at the input of a network. They update the input by minimizing a cost function defined between the generated style image and the real image. They apply gradient descent method not to the weights of the model, but to the random noise for updating the adversarial image. 

 On the contrary, in our suggested method, we start from an original image $x$, which can be correctly labeled by a model $M$ and update it until the network notices the difference between the original image and the distorted image. Formally speaking, we distort $x(k)$ until  $M[x(k)] \neq  y $. 

At each iteration, the input image is distorted by the gradient descent method to generate a \textit{slightly} more noisy image, $x(k)$, until  the model outputs a false label.

We generate JND adversarial images for  image classification and object detection tasks. For the object detection task, we use two attack approaches, i) targeted attacks and ii) non-targeted attacks. In the targeted attack, we generate an adversarial example with an adversarial label. In this approach, an image with an adversarial object label is given at the input and the model is tricked to detect an adversarial label. For example, an image without the car object fools the network, so that it generates a car label at the output. In the non-targeted attack, we generate an adversarial example, where the model assigns a false label to the image. For example, an image with a car fools the network as if there is no car in the image.

\subsection{Cost Function }

The most crucial part of the suggested method is defining a cost function, which enforces the generated adversarial image to be  perceptually similar to the true image while minimizing the error  between the true label and assigned label. In order to achieve this task we add three regularization terms to the cost function, as defined below;

\begin{align}
    \label{equ:cost}
    \mathrm{Cost}(x(k)) = \lambda_1 \mathrm{Loss}(\hat{y}, y) + \lambda_2  ||x(k) - x(0) ||_2^2 \nonumber + \\ 
    \lambda_3 \mathrm{BR}(x(k)) + \lambda_4 \mathrm{TV}(x(k)),
\end{align}

\noindent where $x(0)$ is the input image, $x(k)$ is the updated adversarial image at iteration k, $\hat{y}$ is the output of the model given the updated adversarial image, $x(k)$ and $y$ is the true output of the model. The first term is the loss function. The definition of the  Loss depends on the model we are tricking. For example, if the task is image classification, we can use the cross-entropy loss. 
The second term of the cost of function, $\lambda_2 ||x(k) - x(0) ||_2^2$,  is the $L_2$ distance between the perturbed image and the original image. Interestingly, the adversarial images diverge from the original image, when we use distances other then $L_2$. Note that, when the model $M$ is tricked by the generated adversarial image for the first time at iteration $K$, this term becomes the JND, defined for $L_2$ norm. 

By adding the regularization terms, $\mathrm{BR}(x(k))$ and $\mathrm{TV}(x(k))$, to the cost function, we try to capture the Just Noticeable Different image for the model.  $BR$ is the bounded range loss, which ensures the natural appearance of the adversarial image, whereas $TV$  is the total variation of the image, which enforces color similarity in a neighborhood. The formal definitions of these regularization terms will be provided in the next subsection. The parameters, $\lambda_1, \lambda_2, \lambda_3, \lambda_4$, are experimentally optimized for each training set to establish a balance among the regularization terms.

After we calculate the cost from  Equation \ref{equ:cost}, we update the input image, $x(k)$ by using the gradient descent method, 

\begin{equation}
    \label{equ:update}
    x(k+1) \leftarrow x(k) + n(k),
\end{equation}
\noindent where  $n(k)$ is the  noise added to the image at iteration $k$, defined below;

\begin{equation}
    \label{equ:noise}
    n(k) = - \mathrm{\alpha} \frac{\partial \mathrm{Cost}(x(k))}{\partial x(k)},
\end{equation}
\noindent where $\alpha$ is the learning rate and $Cost$ is the function defined in  Equation \ref{equ:cost}.

Minimizing the cost function iteratively generates a slightly distorted image $x(k)$, at each step.  The image $x(K)$,  generated at the smallest iteration $K$, is the JND adversarial image.

\subsection{Regularizing the Cost Function for Just Noticeable Difference}
\label{regularization_section}

\label{sec:regular}

The third and fourth terms of the cost function, namely, bounded range and total variation,  enforces the natural appearance of the generated adversarial images. Mahendran et al. \cite{mahendran_visualizing_2016} showed that these techniques make a random image look more natural for visualizing the Convolutional Neural Networks.  

\subsubsection{Bounded Range} 
We employ the bounded range loss to  regularize the cost function of the model, where each pixel is penalized if it has an intensity that is not in the range of the  natural image.

Loosely speaking, the bounded range loss penalizes the pixels with intensity values outside the range of the original image. For example, an 8 bits/pixel image is forced to stay in the range of 0 to 255. Since this condition is not guaranteed in the original formulation \cite{mahendran_visualizing_2016},  we further clamp the outlier pixel intensity values. Formally speaking, we define the bounded range loss as follows;

\begin{align}
    \mathrm{BR}(p) = \begin{cases} 
      -p & p \le 0 \\
      p - 255 & p \ge 255 \\
      0 & \mathrm{otherwise,}
   \end{cases}
\end{align}

\noindent where $p$ is the intensity of the  pixel. This function is applied to each pixel of the image to assure that all the pixel values of the image are in the range of [0 - 255]. We normalize the BR loss by dividing it by the number of pixels.

\subsubsection{Total Variation} 
In natural images, a pixel has a ``similar" intensity value with its neighbors. As it is stated by Mahendran et al. \cite{mahendran_visualizing_2016}, this property can be partially simulated by the total variation regularization technique. In this technique, a pixel is penalized if it does not have a ``similar" intensity with its neighbors. Therefore, we penalize the image to force all pixels to have similar intensity values with their neighbors. We use the below formula to penalize the variation among pixels;

\begin{align}
    \mathrm{TV} = \frac{1}{HWC} \sum\limits_{uvk} [(x(v,u + 1,k) - \nonumber
    x(v,u,k))^2 + \\ (x(v+1, u , k) - x(v, u, k) )^2] ,
\end{align}
\noindent where $H, W, C$ are height, width, and the number of channels respectively, u,v,k are values to iterate over the dimensions height, width, and depth. We normalize the TV loss by dividing it by the number of pixels.

Note that, in  Equation \ref{equ:update}, the noise update rule is defined as the partial derivative of the cost function, which distorts the image by the least amount of variation, at each iteration. Due to the bounded range and total variation regularization, included in the cost function, the generated  images are expected to be relatively more consistent to the human perception, compared to the cost functions, which exclude these regularizations. The image, $x(K)$, generated at iteration $K$, where the machine starts to detect the change for the first time, resembles  just noticeably different image perceived by a human.  Thus, it is called JND image perceived by the model, $M$.

\section{Experimental Results}

In this section, we demonstrate the power of the suggested JND method for generating adversarial images and for tricking the classifiers and object detectors.  We compare the JND method with the well-known methods, namely, Fast Gradient Sign  (FSGM)  \cite{goodfellow2014explaining}, Fast Gradient Value (FGV)  \cite{rozsa_adversarial_2016} and DeepFool   \cite{moosavi-dezfooli_deepfool_2016} methods, and the baseline method which correspond to JND method without regularization terms. Without regularization terms, the baseline method cannot preserve the naturalness of the image and produces noisy images.

In order to compare the quality of the adversarial images generated by the above-mentioned methods, we measure several image quality metrics, namely, Peak Signal to Noise Ratio (PSNR),  Structural Similarity (SSIM), Universal Image Quality Index (UQI),  Spatial Correlation Coefficient (SCC) and  Visual Information Fidelity (VIFP). We also compare the average distance between the original and adversarial images using $L_1$, $L_2$, and $L_\infty$ norms.

We observe the effect of the regularization terms on the quality of the generated adversarial images by comparing the JND method with a  baseline method, which omits the total variation and bounded range regularization terms.

We conduct experiments for image classification and object detection tasks:
\begin{itemize}
    \item For  image classification, we generate adversarial images to trick a pre-trained Inception v3 image classifier \cite{szegedy_rethinking_2016} on ImageNet dataset and a pre-trained VGG16 classifier \cite{simonyan2014very} on CIFAR10 dataset.
    \item For  object detection,  we generate  adversarial images to trick a pre-trained RetinaNet object detector \cite{lin_focal_2017} on MS COCO dataset \cite{lin_microsoft_2014}.
\end{itemize}

  



 Finally, we analyze the statistical similarities of the original images to the  adversarial images using  two different measures:
\begin{itemize}
   
    \item Kullback-Leibler distances between the color distributions of the original and adversarial images and  
    \item  $L2$ distances between the original images and the adversarial images.
  \end{itemize} 
  
In order to find optimal parameters for each method, we conduct hyperparameter optimization using grid search. We create a held-out validation set, and select parameters that maximize $\frac{\mathrm{PSNR}*\mathrm{SSIM}}{\mathrm{num\_iters}}$, where PSNR and SSIM are Peak Signal to Noise Ratio, \cite{huynh-thu_scope_2008}, and Structural Similarity, \cite{wang2004image} of the generated adversarial image, and num\_iters is the number of iterations to generate an adversarial image. In this way, we select the hyperparameters that simultaneously maximize the quality of the generated image, PSNR and SSIM, and minimize the number of iterations to generate an adversarial image.

\subsection{ Experiments on CIFAR10 Dataset}
\label{sec:cifar_exp}

In this section, first, we conduct comparative quality analyses of the adversarial images generated from the CIFAR10  dataset using several image quality metrics suggested in previous work \cite{huynh-thu_scope_2008, wang2004image, wang_universal_2002, wald_quality_2000,sheikh2006image}. Second, we compare the distance between the original images and the generated adversarial images. Finally, we compare the classifier attacks of the JND method with Fast Gradient Sign Method (FSGM) of \cite{goodfellow2014explaining} and Fast Gradient Value method (FGV) of \cite{rozsa_adversarial_2016}.



We chose the $\lambda$ values in Equation \ref{equ:cost}, epsilon values for JND, FGSM, and FGV by a hyperparameter search, such that the methods generate the ``best quality" images with a small iteration number. In this regard, we maximize  Peak Signal to Noise Ratio (PSNR) \cite{huynh-thu_scope_2008} and Structural Similarity Measure (SSIM) \cite{wang2004image} while minimizing the average number of iterations to generate adversarial images from a subset of images from the training set. 

  We estimate the optimal JND parameters, for the cost function as   $\lambda_1 =10, \lambda_2 = 1, \lambda_3 =1, \lambda_4 = 10$  and for the learning rate as $\alpha =0.05$. For FGSM and FGV, the best epsilon value is $0.5$ and $0.4$, respectively. For DeepFool, we used the default parameters in the paper. 

\subsubsection{Comparative Quality Analyses of the Adversarial Images Generated from  CIFAR10}

Let us start by analyzing the effect of the TV (total variation) and BR (bounded range)  regularization on the quality of the adversarial images. Figure \ref{fig:cifar10_images} shows sample images generated by the JND  and the baseline method, where we omit the regularization terms in the cost function of  Equation \ref{equ:cost}. Due to the low resolution of images ($32\times32$) in the CIFAR10 dataset, none of the images look natural at all.  However, visual inspection reveals that the JND method generates images closer to the original images compared to that of the baseline method. 
\begin{figure}[h]
    \centering
    \begin{minipage}{0.48\textwidth}
        \centering
        \includegraphics[width=\textwidth]{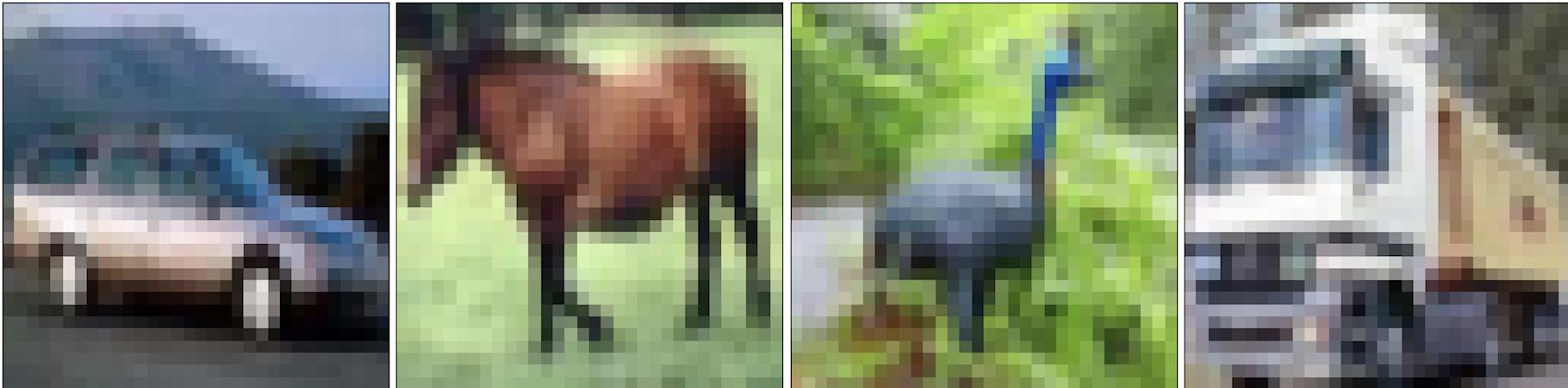}
        \captionsetup{labelformat=empty}
        \caption{Original images}
    \end{minipage}\hfill
    \begin{minipage}{0.48\textwidth}
        \centering
        \includegraphics[width=\textwidth]{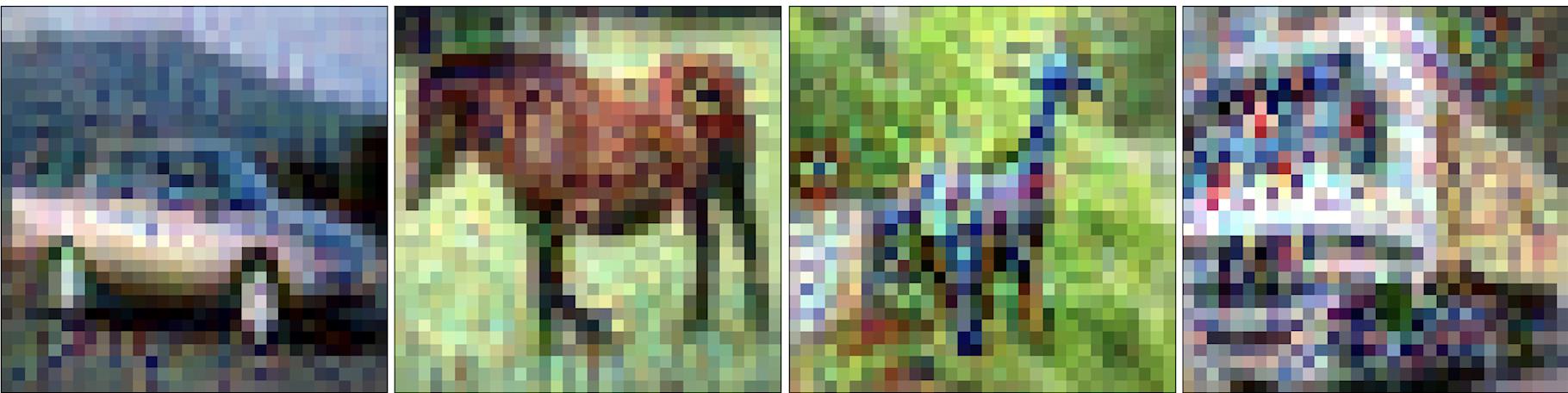}
        \captionsetup{labelformat=empty}
        \caption{JND images (with regularization)}
    \end{minipage}\hfill
    \begin{minipage}{0.48\textwidth}
        \centering
        \includegraphics[width=\textwidth]{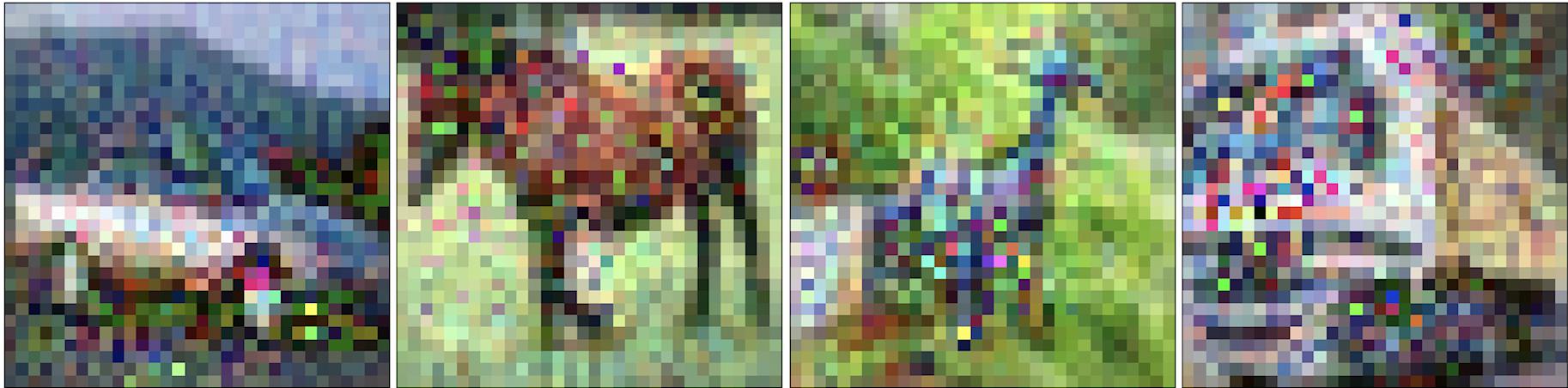}
        \captionsetup{labelformat=empty}
        \vspace{-5mm}
        \caption{Baseline images (without regularization)}
    \end{minipage}
\setcounter{figure}{0}
\caption{Sample adversarial images generated by the JND method and the baseline method, where we omit the total variation and bounded range regularizations: Top row: Original images taken from CIFAR10 dataset. Middle row: JND adversarial images, generated with regularization. Bottom row: Adversarial images generated by baseline method without any regularization.}
\label{fig:cifar10_images}
\end{figure}

In the next set of experiments, we compare the quality of the adversarial images generated by JND,  FSGM, FGV, DeepFool, and the baseline methods, where we neglect the regularization terms. We report the image quality assessments for the generated images in Table \ref{table:quality_cifar}. The images generated by the JND method outperform all methods with respect to the image quality measures. This quantifies the effect of the regularization on the image quality of adversarial images. Note that the JND method substantially outperforms other methods in the PSNR metric. The rest of the quality measures are very similar for the images generated by JND and the other methods.  This result reveals that the low-resolution images in the CIFAR10 dataset bring an upper bound to the quality of the adversarial images. Thus, we may conclude that the resolution of images in the dataset is an important factor for the generation of high-quality adversarial images.

All of the generated adversarial images with and without regularization successfully tricked the classifier.

\begin{table}[h]
\centering
\caption{Image quality metrics on generated adversarial images by the baseline  (without the BR and TV regularization terms), suggested JND, FGSM, FGV and DeepFool methods on CIFAR10 dataset.  The metrics are   Peak Signal to Noise Ratio (PSNR),  Structural Similarity (SSIM), Universal Image Quality Index (UQI),  Spatial Correlation Coefficient (SCC) and  Visual Information Fidelity (VIFP). For all metrics, the higher is the better. We report the average values of the metrics calculated over 300 generated examples in the CIFAR10 dataset.}
\label{table:quality_cifar}
\setlength\tabcolsep{2pt}
\begin{tabular}{lccccc}
\hline
Metrics &    Baseline & JND & FGSM & FGV & DeepFool \\ \hline
PSNR \cite{huynh-thu_scope_2008}       & 338 &  \textbf{382}     & 370 & 371 & 373      \\
SSIM \cite{wang2004image}   & 0.953 &  \textbf{0.998}   & 0.989 & 0.985 & 0.991 \\
UQI \cite{wang_universal_2002}     & 0.952 &  \textbf{0.994}   & 0.990 & 0.987 & 0.992    \\ 
SCC\cite{wald_quality_2000} & 0.917 & \textbf{0.943}      & 0.921 & 0.918 & 0.931      \\ 
VIFP \cite{sheikh2006image}     & 0.984 & \textbf{0.998}      & 0.994 & 0.991 & 0.996       \\ \hline
\end{tabular}
\end{table}

We also compare the suggested JND method with FGSM, FGV, DeepFool, and the baseline methods, in terms of  $L_{1}$, $L_{2}$ and $L_{\infty}$  distances between the original and adversarial images generated from the CIFAR10 dataset. We report the average distances, estimated over 300 images, in Table \ref{table:cifar_deepfool}. Note that the suggested JND method generates closer adversarial images to the original ones relative to the rest of the methods.

Since the CIFAR10 dataset contains 10 classes, the speed of the generation is not an issue for DeepFool, and the speed difference between both algorithms is negligible. 

\begin{table}[h]
\centering
\caption{The average distances between the original images and the adversarial examples with JND, FGSM, FGV and DeepFool methods generated from CIFAR10 dataset.The averages are estimated over 300 images.  Lower is better for all of the metrics.}
\label{table:cifar_deepfool}
\setlength\tabcolsep{2pt}
\begin{tabular}{lcccc}
\hline
Metrics   & JND & FGSM & FGV  & DeepFool\\ \hline
$L_1$       & \textbf{11.64}       & 33.4 & 42.11 & 24.08      \\ 
$L_2$    & \textbf{0.21}      & 0.70 & 1.12 & 0.54        \\ 
$L_{\infty}$ & \textbf{0.004}  &  0.020 & 0.082 & 0.008   \\ 
\hline
\end{tabular}
\end{table}

\subsubsection{Classifier attacks on CIFAR10}

We conduct a set of experiments to compare the convergence rates of our JND method with FSGM   and FGV methods while the adversarial images attack a classifier. Recall that FGSM  employs the sign of the gradient while updating the true image in a single step, whereas the FGV method employs the gradient value itself for updating the true image in a single step. Their cost function is different from the suggested JND cost, which involves additional BR and TV regularization terms. The single-step adversarial image computation of FSGM is given in equation \ref{eq:fgsm}:

\begin{equation}
    \label{eq:fgsm}
    x_{FGSM}^{adv} = x + \epsilon\  \mathrm{sign}(\nabla_x L(h(x)), y_{true}) ,
\end{equation}
\noindent where $h$ is the target model, $x$ is the image and $y_{true}$ is the true label. Originally, the FGSM attack is a single step attack. However, applying this method iteratively to the target model yields higher error rates as stated by Kurakin et al. \cite{kurakin_adversarial_2016}. 

The single-step adversarial image computation of FGV is given in equation \ref{eq:fgv}:

\begin{equation}
    \label{eq:fgv}
    x_{FGV}^{adv} = x + \epsilon \nabla_x L(h(x), y_{true}) ,
\end{equation}

\noindent where $h$ is the target model, $x$ is the image, $y_{true}$ is the true label. Originally, the FGV attack is a single-step attack. However, we applied this method iteratively as in the FGSM method for a fair comparison. 

We compare our JND method with the FGSM attack and FGV attack on the CIFAR10 dataset. Since the DeepFool method checks other classes to find the minimum perturbation, it is incompatible with JND, FGSM, and FGV methods. Thus, we suffice to conduct this experiment with only FGSM and FGV methods. We generate adversarial images, which trick the classifier with the given confidence score value and measure the quality of the images and speed of the algorithm. We use two different confidence scores, $25\%$ and $60\%$, and measure the average number of iteration to reach the confidence score, SSIM scores, and $L_2$ distances between original and generated images. The results, reported in Table \ref{table:cifar_attacks} indicate that the JND method generates the most natural images (SSIM) and the closest images ($L_2$) to the original ones when the classifier is tricked, yet it is the fastest method, which converges to the given confidence score value.

\begin{table}[h]
\centering
\caption {Similarity measurement, when the model is tricked with the given confidence value for the CIFAR10 dataset. The values are calculated by averaging 500 images from the test set. Each model has the best hyperparameters, which were searched on the training set before. Higher is better for SSIM, and lower is better for $L_2$ and the number of iterations.} 
\label{table:cifar_attacks}
\begin{tabular}{lcccc}
 Method & Confidence & Iterations & SSIM & $L_2$    \\    \hline
 FGSM  & 0.25 &  126.1   & 0.9962 & 0.393 \\
 FGV   & 0.25 &  269.2   & 0.9902 & 0.236 \\
 JND   & 0.25 &  \textbf{91.3}   & \textbf{0.9981} & \textbf{0.163} \\
 \hline
 FGSM  & 0.60 &  143.5   & 0.9935 & 0.403 \\
 FGV   & 0.60 &  294.8   & 0.9892 & 0.458 \\
 JND   & 0.60 &  \textbf{125.2}   & \textbf{0.9969} & \textbf{0.265} \\
 \hline
\end{tabular}
\end{table}


While conducting the experiments, we select the best hyperparameters for all methods with the grid search. The attacked model is VGG16 \cite{simonyan2014very} and its accuracy without any adversarial attacks is 88\%.

\begin{figure}[h!]
    \centering
\begin{minipage}{\linewidth}
\centering
\end{minipage}
\begin{minipage}{\linewidth}
\centering
\begin{minipage}{0.24\linewidth}
\includegraphics[width=\textwidth]{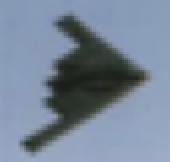}
\end{minipage}
\hfill
\begin{minipage}{0.24\linewidth}
\includegraphics[width=\textwidth]{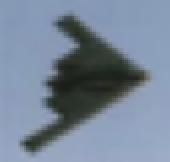}
\end{minipage}
\hfill
\begin{minipage}{0.24\linewidth}
\includegraphics[width=\textwidth]{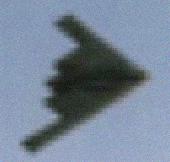}
\end{minipage}
\hfill
\begin{minipage}{0.24\linewidth}
\includegraphics[width=\textwidth]{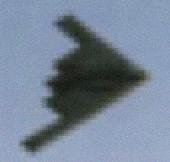}
\end{minipage}
\vspace{1mm}
\end{minipage}

\begin{minipage}{\linewidth}
\begin{minipage}{0.24\linewidth}
\includegraphics[width=\textwidth]{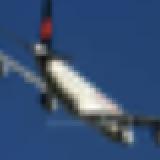}
\captionsetup{labelformat=empty}
\caption{Original images}
\end{minipage}
\hfill
\begin{minipage}{0.24\linewidth}
\includegraphics[width=\textwidth]{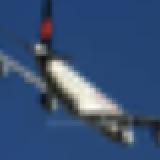}
\captionsetup{labelformat=empty}
\caption{JND images \cite{9191090}}
\end{minipage}
\hfill
\begin{minipage}{0.24\linewidth}
\includegraphics[width=\textwidth]{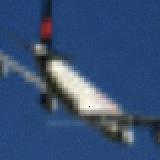}
\captionsetup{labelformat=empty}
\caption{FGSM images \cite{goodfellow2014explaining}}
\end{minipage}
\hfill
\begin{minipage}{0.24\linewidth}
\includegraphics[width=\textwidth]{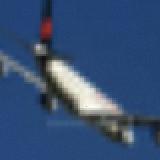}
\captionsetup{labelformat=empty}
\caption{FGV images \cite{rozsa_adversarial_2016}}
\end{minipage}

\end{minipage}
    \setcounter{figure}{1}
    \caption{Generated adversarial images, when conducting attacks on the CIFAR10 dataset. 
    Top: Images, when the model is tricked with at least 25\% confidence scores. Bottom:Images, when the model is tricked with at least 60\% confidence scores. All of the SSIM scores of the generated images are at the level of 0.99. However, it can be seen in Table \ref{table:cifar_attacks} that JND images are closer to the original images compared to FSGM and FGV images.}
\label{fig:cifar10_attack_images}
\vspace{-5mm}
\end{figure}

In Figure \ref{fig:cifar10_attack_images}, we show sample adversarial images generated by FSGM, FGV and JND methods. Due to the low resolution of the CIFAR10 dataset, the adversarial images look visually similar. However, analyzing Table \ref{table:cifar_attacks} shows that although the  SSIM scores are very high for all the methods, the measured image qualities are relatively higher for the JND method. 

\subsection{ Experiments on ImageNet Dataset}
\label{sec:imagenet_exp}

In this section,  we repeat the image quality analysis and classifier attack experiments of the previous section on the ImageNet dataset. 


 We choose the best hyperparameters that maximize the PSNR and SSIM scores and minimize the average number of iterations in a selected sample set. For JND parameters, the best values turn out to be $100,10,10,1$ for  $\lambda_1, \lambda_2, \lambda_3, \lambda_4$, respectively, and $0.0001$ as learning rate $\alpha$ value. For FGSM and FGV, the best epsilon value is $0.1$. For DeepFool, we used the default parameters in the paper.

\subsubsection{Comparative Quality Analyses of the Adversarial Images Generated from  ImageNet Dataset}
In order to observe the effect of the regularization on the quality of the adversarial images, we generate two sets of adversarial images, which can successfully trick image classifiers.  The first set of images are generated by the baseline method,  without  TV and BR regularization terms. The second set consists of the JND adversarial images generated with all the regularization terms, suggested in Equation \ref{equ:cost}.

\begin{figure}[h!]
\centering
\begin{subfigure}[b]{0.49\textwidth}
   \includegraphics[width=1\linewidth]{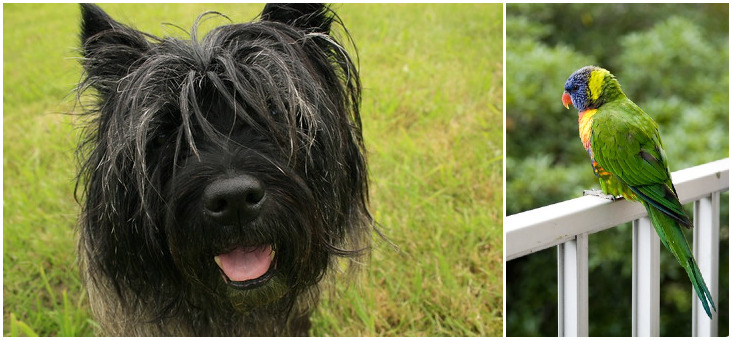}
   \captionsetup{labelformat=empty}
   \caption{Original images}
\end{subfigure}

\begin{subfigure}[b]{0.49\textwidth}
   \includegraphics[width=1\linewidth]{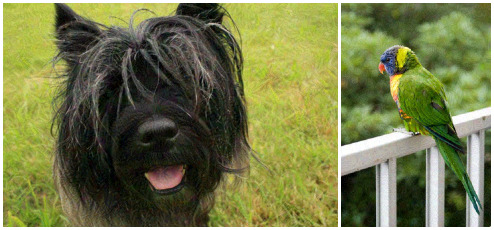}
   \captionsetup{labelformat=empty}
   \caption{JND images (with regularization)}
\end{subfigure}
\begin{subfigure}[b]{0.49\textwidth}
   \includegraphics[width=1\linewidth]{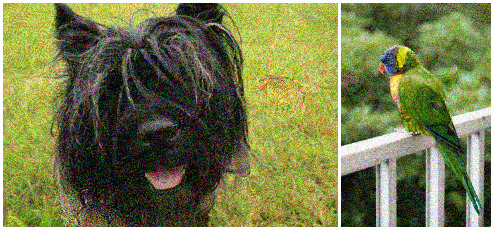}
   \captionsetup{labelformat=empty}
   \caption{Baseline images (without regularization)}
\end{subfigure}
\caption{Adversarial images generated from ImageNet dataset: First row: Samples of original images. Middle row: Just Noticeably Different Adversarial Images with using regularization functions at the time when they first trick the network. Last row: Samples generated without any regularization. All the samples trick classifiers successfully.}
\label{fig:imagenet_images}
\end{figure}

Figure \ref{fig:imagenet_images} shows two sample images from ImageNet. Note that, JND method increases the naturalness of the generated adversarial images, indicating the importance of the regularization with TV and BR, in high-resolution images ($224\times224$) of ImageNet.

\begin{figure}[h!]
    \centering
    \includegraphics[width=\linewidth]{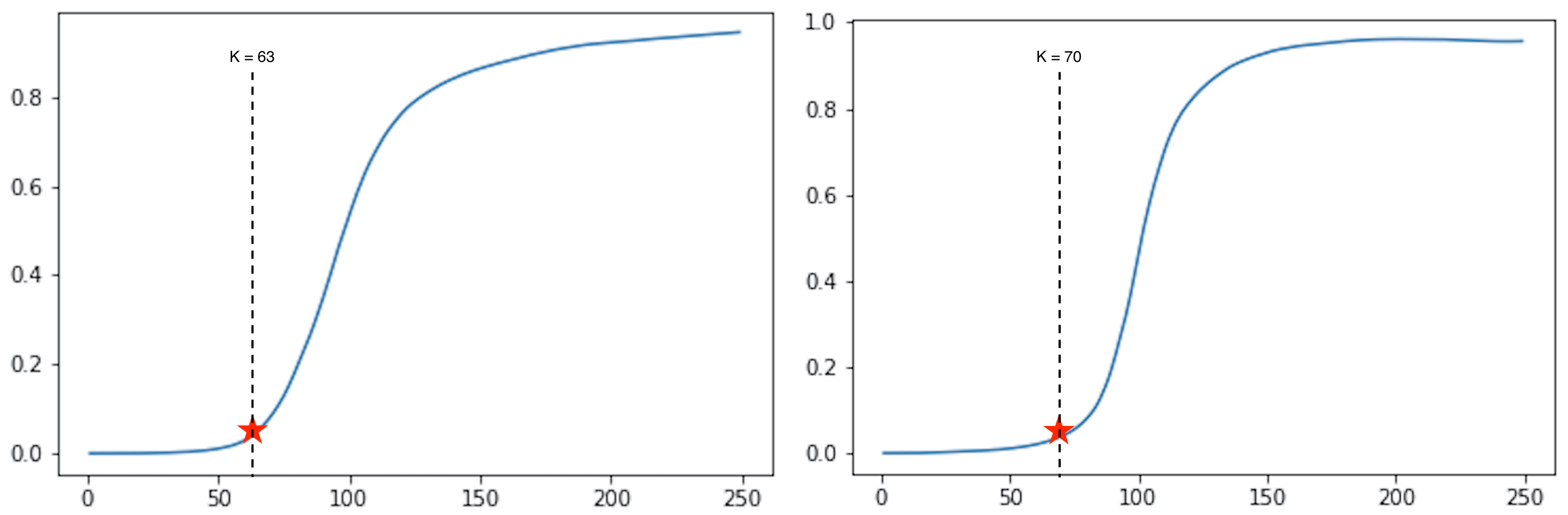}
    \caption{The confidence scores of the updated dog (left) and parrot (right) images. After K=63 and 70, respectively, the images successfully trick the network. Therefore, Just Noticeably Different Adversarial image is generated at K=63 and 70, respectively. Y-axis shows the confidence scores for the target class, in 0-1 scale and the X-axis shows the number of iterations.}
    \label{fig:conf_scores}
\end{figure}

In Table \ref{table:imagenet-quality}, we report the image quality measures averaged over 300 generated examples. The quality of the images is the best for the JND method compared to the adversarial images generated by the baseline,  FGSM \cite{goodfellow2014explaining}, FGV \cite{rozsa_adversarial_2016} and DeepFool methods \cite{moosavi-dezfooli_deepfool_2016}.

\begin{table}[h]
\centering
\caption{Image quality metrics on generated adversarial images by the baseline method (without the regularization terms), suggested JND method, FGSM method, FGV method, and DeepFool method on ImageNet dataset.  The metrics are Peak Signal to Noise Ratio (PSNR),  Structural Similarity (SSIM),  Universal Image Quality Index (UQI),  Spatial Correlation Coefficient (SCC) and  Visual Information Fidelity (VIFP). For all metrics, the higher is, the better. We report the average values of the metrics calculated over 300 generated examples in the ImageNet dataset.}
\label{table:imagenet-quality}
\begin{tabular}{lccccc}
\hline
Metrics &    Baseline & JND & FGSM & FGV & DeepFool \\ \hline
PSNR \cite{huynh-thu_scope_2008}       & 350 &  \textbf{392}     & 382 & 384 & 388      \\
SSIM \cite{wang2004image}   & 0.963 &  \textbf{0.9998}   & 0.9984 & 0.9972 & 0.9993 \\
UQI \cite{wang_universal_2002}     & 0.962 &  \textbf{0.9998}   & 0.9976 & 0.9981 & 0.9991    \\ 
SCC \cite{wald_quality_2000} & 0.9172 & \textbf{0.9940}      & 0.9434 & 0.9510 & 0.9912     \\ 
VIFP \cite{sheikh2006image}     & 0.9842 & \textbf{0.9998}      & 0.9981 & 0.9991 & 0.9995       \\ \hline
\end{tabular}
\end{table}

We compare the distance between the original images and the corresponding adversarial images using  $L_1, L_2$ and $L_{\infty}$ norms in the ImageNet dataset.
Our method generates closer adversarial images to the original ones in terms of $L_{\infty}$ norm, which is a widely used metric to make comparisons. However, DeepFool generates closer adversarial images in terms of $L_1$ and $L_2$ norms. We report the distances between the original images and the generated adversarial images in Table \ref{table:imagenet_deepfool}.

Our method also outperforms DeepFool in terms of execution time. DeepFool checks all the adversarial perturbations for each class other than the true class to find the minimum perturbation, which is $999$ on the ImageNet dataset. Thus, finding the closest adversarial example to the original one requires nearly $1000$ times more gradient calculations compared to our JND method.

\begin{table}[h]
\centering
\caption{The distances between the original images and the generated adversarial examples with DeepFool \cite{moosavi-dezfooli_deepfool_2016} and JND on ImageNet dataset. Lower is better for all of the metrics.}
\label{table:imagenet_deepfool}
\begin{tabular}{lcccc}
\hline
Metrics   & JND & FGSM & FGV  & DeepFool\\ \hline
$L_1$       & 41.09       & 400.11 & 239.94 & \textbf{33.50}      \\ 
$L_2$    & 0.1185      & 1.0012 & 0.8493 & \textbf{0.1178}        \\ 
$L_{inf}$ & \textbf{0.0002}  &  0.0131 & 0.0338 & 0.0042   \\ 
\hline
\end{tabular}
\end{table}

\begin{figure*}[ht!]
\centering
\begin{minipage}[t]{0.323\linewidth}
\includegraphics[width=\textwidth]{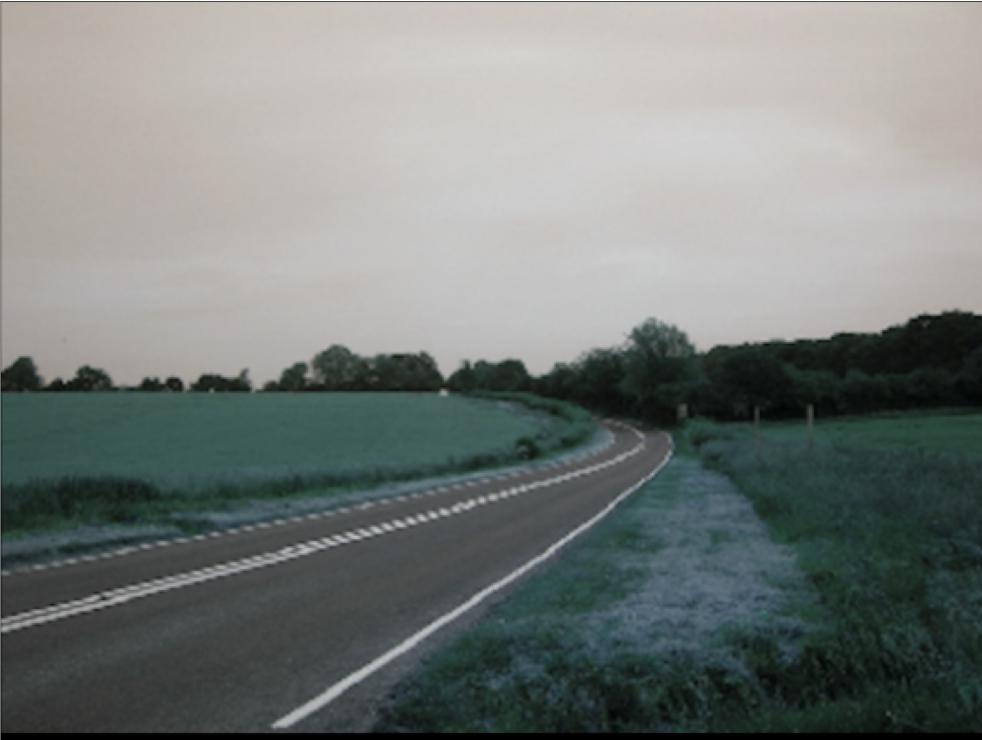}
\captionsetup{labelformat=empty}
\caption{Original image}
\end{minipage}
\begin{minipage}[t]{0.323\linewidth}
\includegraphics[width=\textwidth]{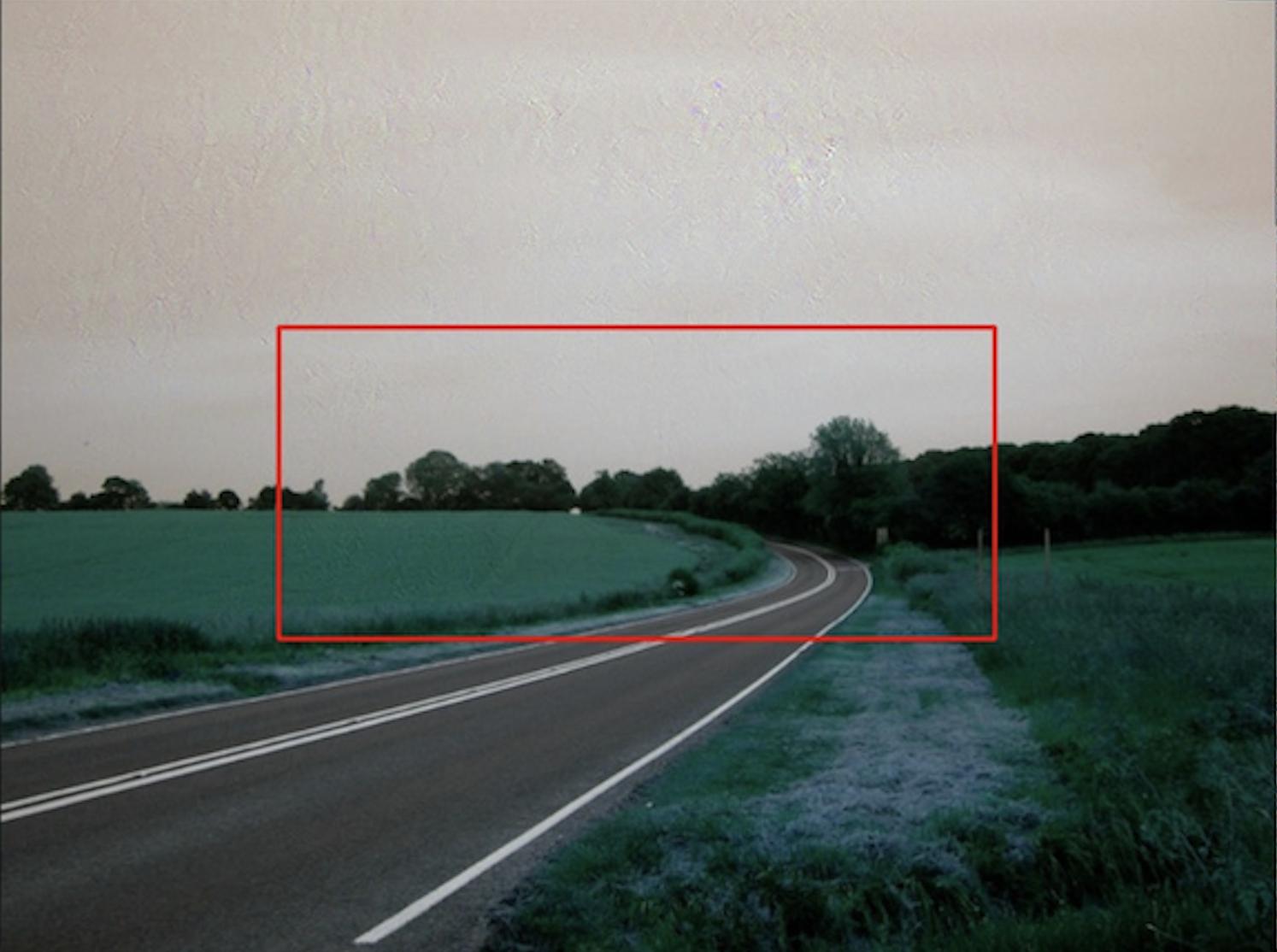}
\captionsetup{labelformat=empty}
\caption{JND image (with regularization)}
\end{minipage}
\begin{minipage}[t]{0.323\linewidth}
\includegraphics[width=\textwidth]{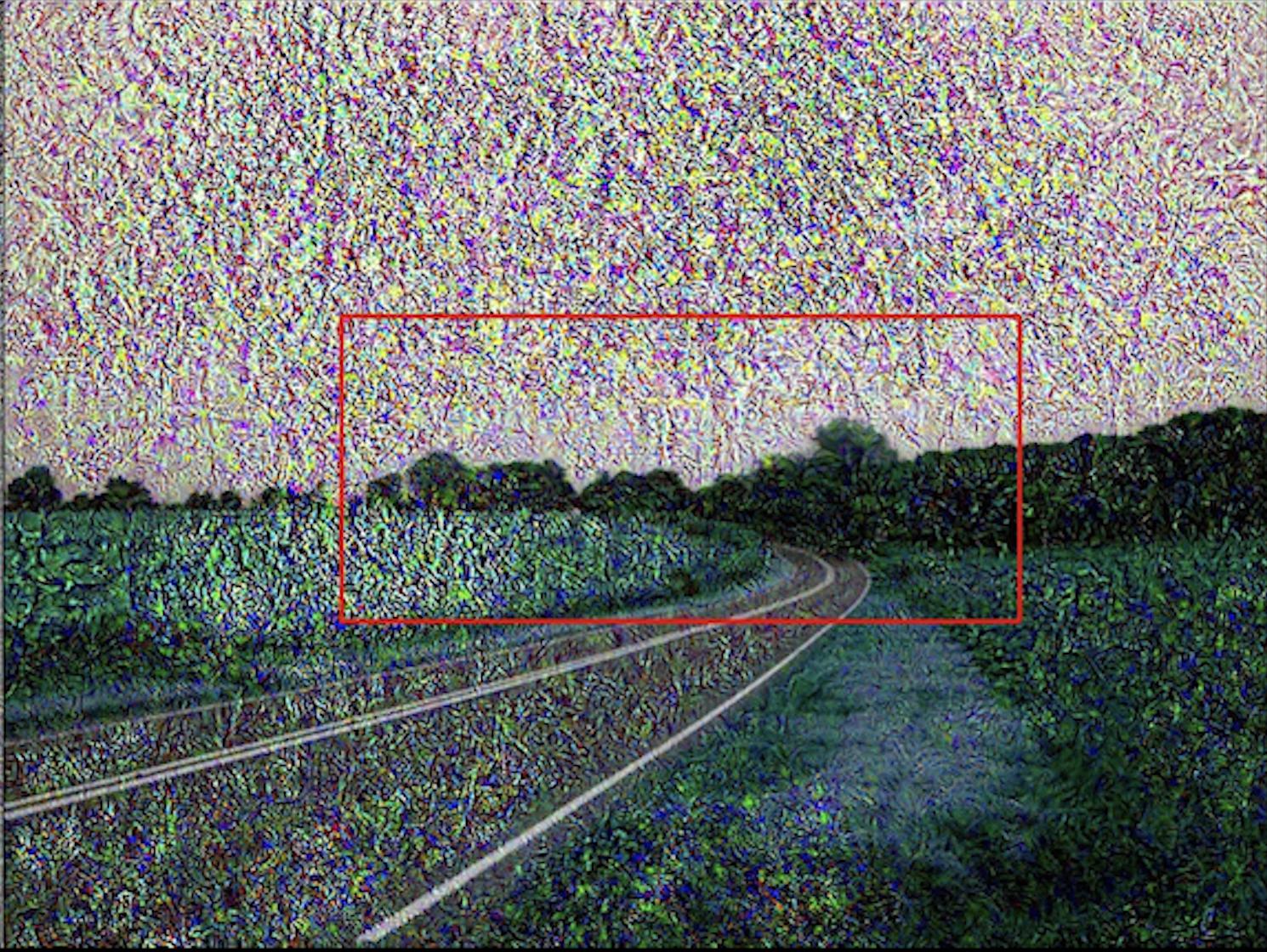}
\captionsetup{labelformat=empty}
\caption{Baseline image (without regularization)}
\end{minipage}
\setcounter{figure}{4}
\caption{Targeted Attacks: Generated images  trick RetinaNet object detector. From left to right: Original image, Just Noticeably Different image, Baseline image (without regularization)}
    \label{fig:obj_detect}
\end{figure*}

\subsubsection{Classifier attacks on ImageNet}

While increasing the quality of images, the JND method does not decrease the confidence of the classifier. The model wrongly classifies all the generated images with nearly 99\% confidence at the end of the generation process. In Figure \ref{fig:conf_scores}, we report the confidence scores of the generated images. Notice that the JND images are generated at the elbow of the confidence curve, which indicates that after the JND images are generated, the confidence of the machine suddenly increases. While the JND dog image is generated at iteration $K=63$, the JND  parrot image is generated at $K=70$. After the generation of the JND image at iteration K, the rest of the generated adversarial images trick the network successfully.

As it is demonstrated in the above experiments, the JND method is more powerful in generating high-quality adversarial images in ImageNet compared to that of CIFAR10. This improvement can be attributed to the high-resolution images of the ImageNet dataset. 

\subsection{Adversarial Image Generation for Object Detection}
\label{sec:obj_exp}

\begin{figure*}[h]

\centering
\begin{minipage}{0.245\linewidth}
\includegraphics[width=\textwidth]{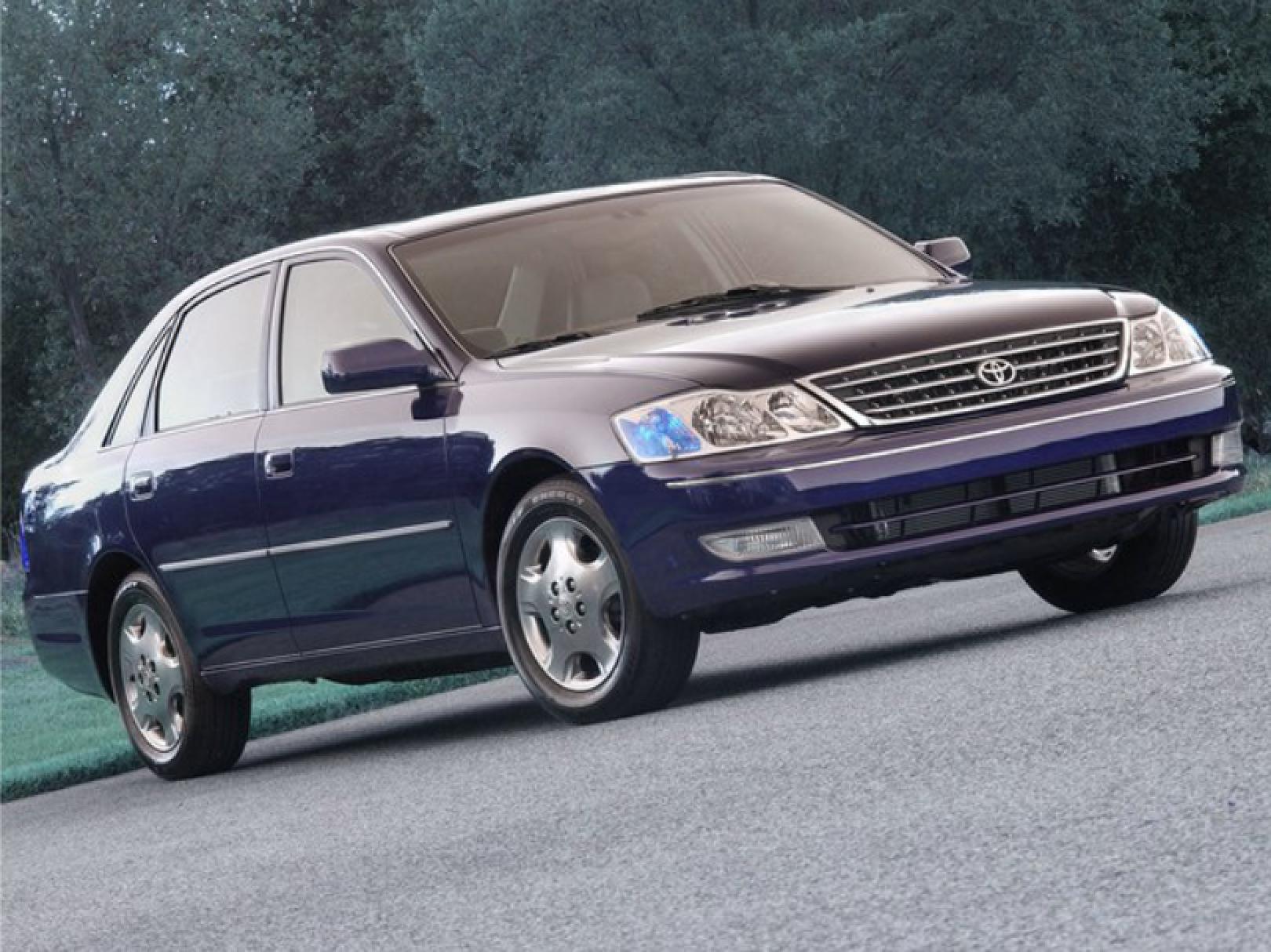}
\captionsetup{labelformat=empty}
\caption{Original image }
\end{minipage}
\hfill
\begin{minipage}{0.245\linewidth}
\includegraphics[width=\textwidth]{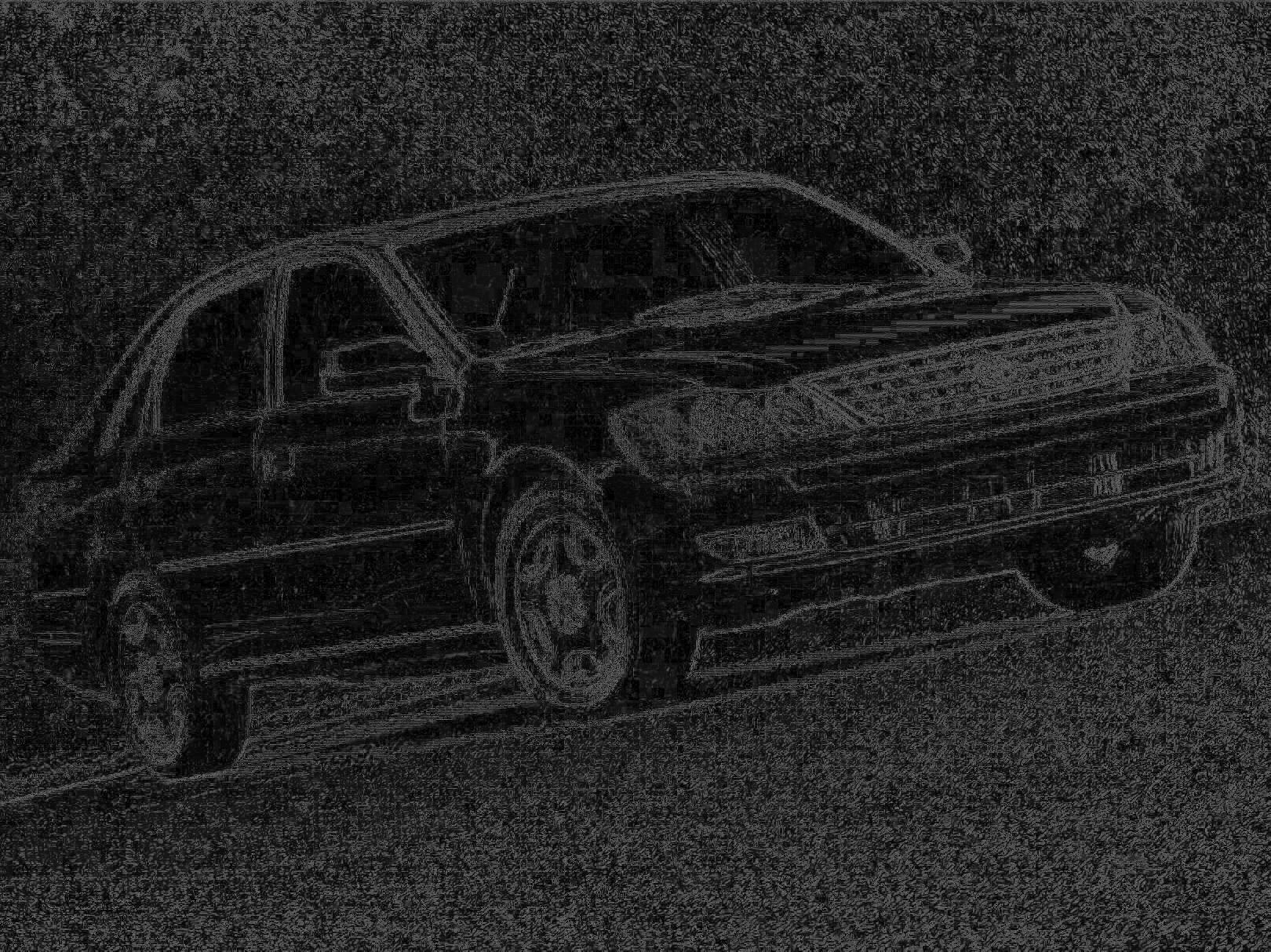}
\captionsetup{labelformat=empty}
\caption{JND  difference}
\end{minipage}
\hfill
\begin{minipage}{0.245\linewidth}
\includegraphics[width=\textwidth]{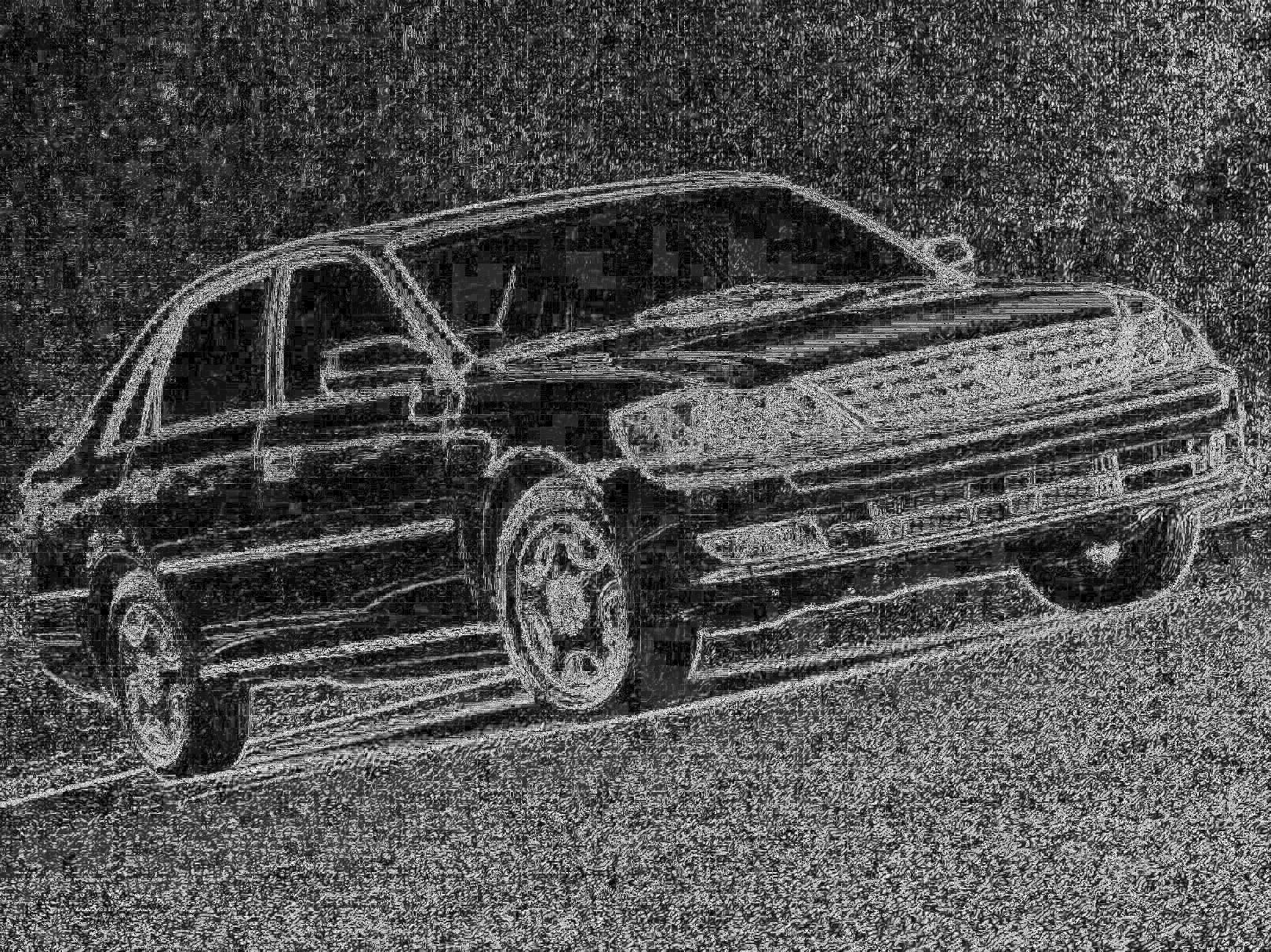}
\captionsetup{labelformat=empty}
\caption{FGSM  difference}
\end{minipage}
\hfill
\begin{minipage}{0.245\linewidth}
\includegraphics[width=\textwidth]{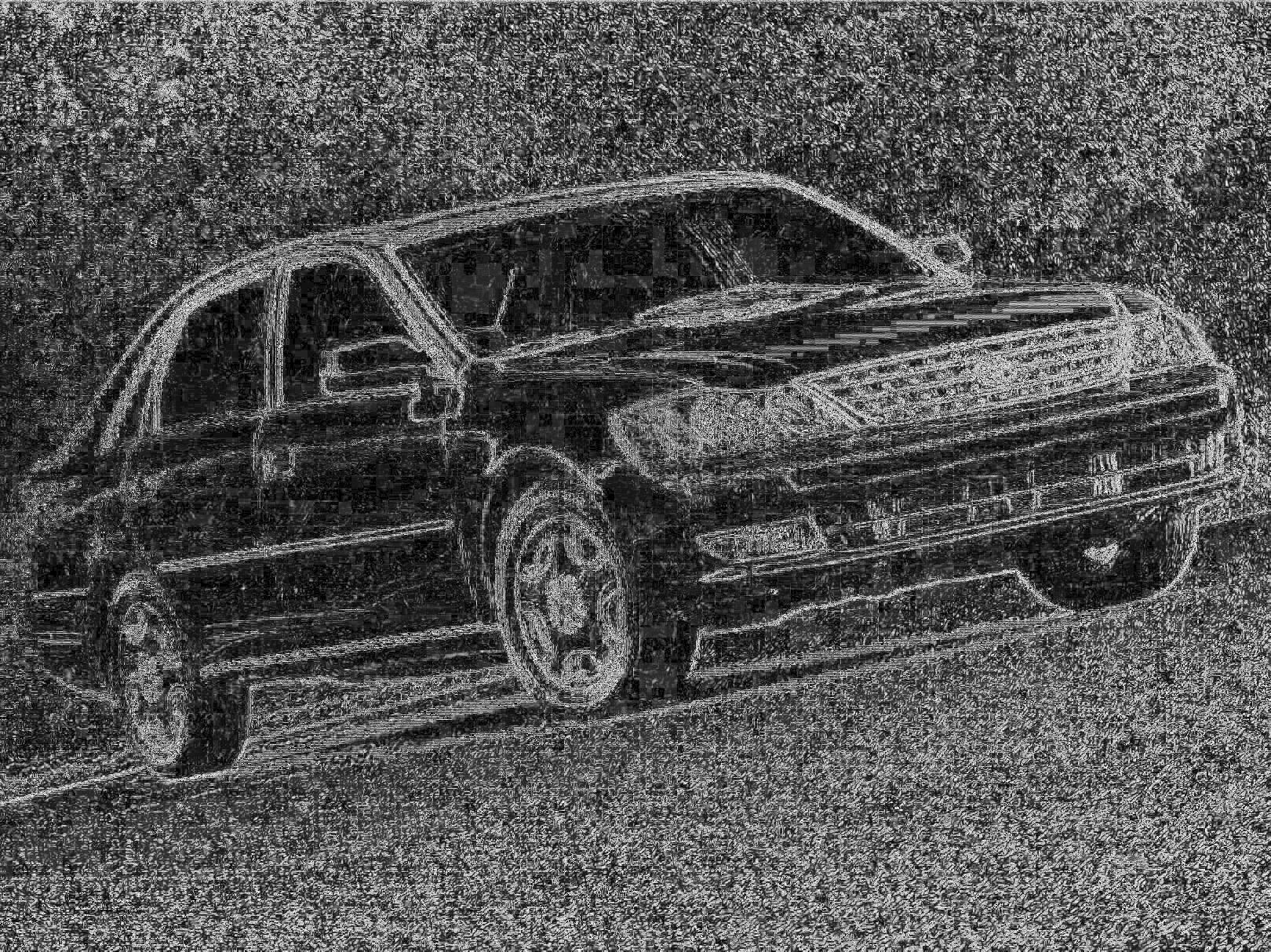}
\captionsetup{labelformat=empty}
\caption{FGV  difference}
\end{minipage}
\setcounter{figure}{5}
\caption{Non-targeted Attacks: Differences from the original image for generated images to trick RetinaNet object detector with non-targeted attacks. From left to right: Original image, difference of the original image with the ``Just Noticeably Different image'', the FGSM image and the FGV image. The model cannot identify the car in the middle of the image. The differences are calculated as $L_2$ norm in channel axis and multiplied by 10 for easy visualization. Colormaps for all three difference images are the same.}
\label{fig:obj_detect_nontargeted}
\end{figure*}

For the object detection task, we use Retina-Net, which is pre-trained on Microsoft COCO \cite{lin_microsoft_2014} dataset. We conduct experiments for targeted and non-targeted attacks. In targeted attacks, we generate an adversarial example with an object that is not present in the original image. Our method successfully generates this type (targeted) of adversarial example while  FGSM and FGV \cite{goodfellow2014explaining} \cite{rozsa_adversarial_2016}, fail to generate adversarial examples since they are non-targeted methods. Both methods are trained for generating adversarial images, which are far from their original label. 

Firstly, we use a simple road image in the Figure \ref{fig:obj_detect}(left) and update it iteratively to generate an adversarial example to trick the classifier in a way that the classifier sees a car in the middle of the image. We generate two different adversarial examples from the road image on the left of Figure \ref{fig:obj_detect}. The image in the middle of Figure \ref{fig:obj_detect} is the Just Noticeably Different adversarial example. The image in the right of Figure \ref{fig:obj_detect} is the baseline image, which is generated without using regularization techniques. As it is stated before, both FGSM and FGV fail to generate this type of adversarial example since they perform only non-targeted attacks.

Secondly, we generate an adversarial example, such that the model cannot see the object, which is actually present in the original image. We use a simple car image and update it iteratively with the methods, JND, FGSM, and FGV. All of the methods successfully generate adversarial examples where the model cannot "see" the cars. We generate three different adversarial examples from the car image, given in Figure \ref{fig:obj_detect_nontargeted}.  Since the chosen epsilon value is too small, all of the images look very similar to each other. However, the closest image to the original one is the Just Noticeably Different image.

The JND method not only generates targeted attacks but also, it can generate non-targeted attacks when given an incorrect label. Moreover, the generated images with JND look more natural and closer to the original images. 

All the $\lambda$ values in Equation \ref{equ:cost} are chosen as $1$ for object detection experiments.

\subsection{Statistical Similarities between the True and Adversarial Images}
\label{sec:stat_exp}

\begin{figure*}[ht]

\centering
\begin{minipage}[t]{0.49\linewidth}
   \includegraphics[width=\linewidth]{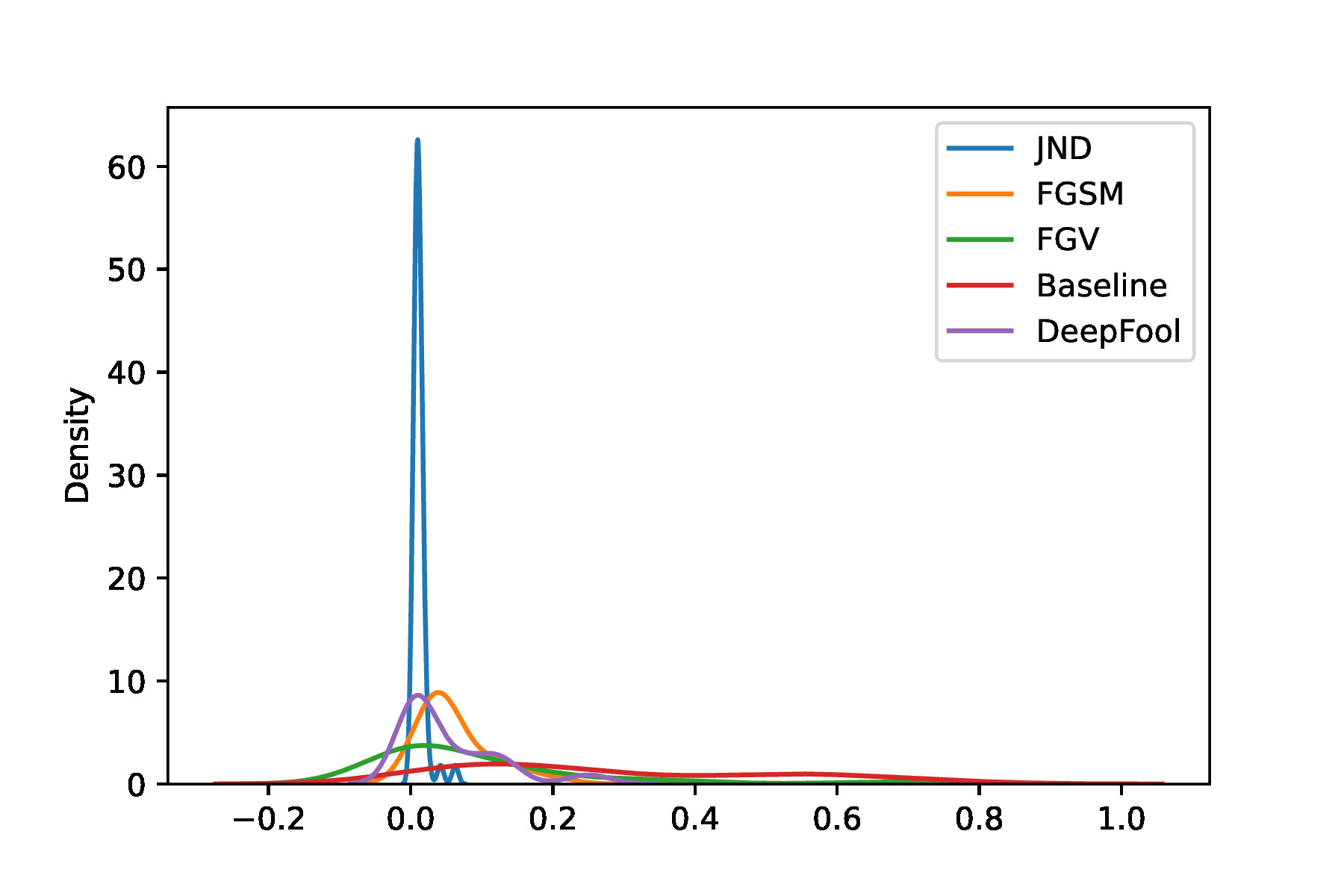}
   \caption{Probability density functions of Kullback-Leibler divergence obtained from CIFAR10 dataset for JND, FSGM, FGV, DeepFool and Baseline methods.}
   \label{fig:kl_dists}
\end{minipage}
\hfill
\begin{minipage}[t]{0.49\linewidth}
\includegraphics[width=\linewidth]{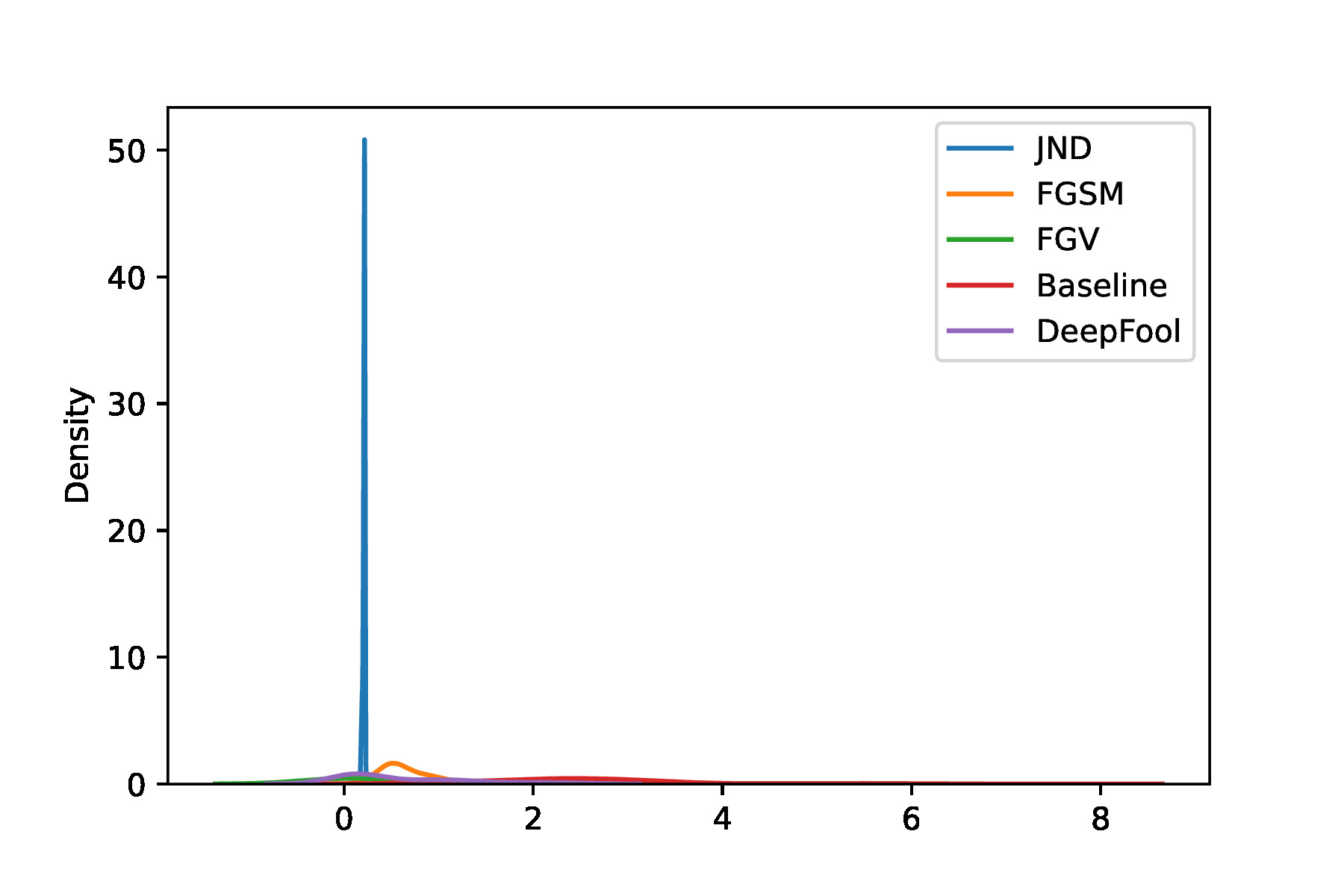}
   \caption{Probability density functions of $L_2$ distance obtained from CIFAR10 dataset for JND, FSGM, FGV, DeepFool and Baseline methods.}
   \label{fig:l2_dists}
\end{minipage}
\end{figure*}

In this section, we analyze the statistical properties of JND images and compare these properties to the adversarial images generated by the baseline (JND without regularization), FSGM, FGV, and DeepFool methods. Since the JND method performs relatively better in the ImageNet dataset, we suffice to make the statistical comparisons on the CIFAR10 dataset.   We  conduct the experiments in two parts: 

\begin{enumerate}


\item We estimate the Kullback-Leibler divergence between the true and adversarial images  for each method by,
\begin{equation}
\label{equ:kl_div}
\mathrm{KL}[p(r)||q(r)] =  - \sum\limits_{x \in X}p(r) \log \frac{p(r)}{q(r)},
\end{equation}
where $p(r)$ and $q(r)$ are the normalized color histograms of the true and adversarial images, respectively. Then, assuming that  $K[p(r)||q(r)]$ is a random variable, for each method, we estimate the probability density function of KL divergences over all images using a kernel density estimation method. Figure \ref{fig:kl_dists} shows the distribution of KL divergences, estimated for the JND, baseline (JND without regularization),  FGSM, FGV, and DeepFool methods. Note that the JND method generates the sharpest distribution around 0 value, which shows that the KL divergence between the adversarial and true images for the JND method is relatively smaller than that of the other methods.

\item 


We  compute $L_2$ distance between the true images and adversarial images for each method by,
\begin{equation}
    \label{equ:l2}
    L_2 = \sum\limits_{ \forall{x_i}} |\hat{x}_i- x_i|^2,
\end{equation}
\noindent where $x_i$ and $\hat{x}_i$ are the pixel color values at $ith$ pixel of the original and adversarial images, respectively. The adversarial images are generated by    JND, FGSM, FGV, DeepFool, and the baseline methods. Then, assuming that $L_2$ is a random variable, for each method, we estimate the probability density function of $L_2$ using a kernel density estimation method, as shown in  Figure \ref{fig:l2_dists}. Note that the JND method has the sharpest distribution of $L_2$ around $0$ value compared to the other methods. This result is consistent with Table \ref{table:cifar_deepfool}, where the JND method gives the smallest $L-$norms between the difference of original and adversarial images.
\end{enumerate}

As it can be observed from Figure  \ref{fig:l2_dists}, the distribution of all methods has almost zero mean, except the baseline method. However, the standard deviations of the methods are relatively large compared to that of the JND method. This result indicates that almost all of the adversarial images generated by the JND method are very similar to the original images, yet they trick the classifiers with very high confidence rates.

\section{Conclusion}
In this study, we adopt the concept of Just Noticeable Difference (JND) to machine perception. We define the JND images for a machine learning model by adding a just discriminating noise to the image, such that the machine is fooled. 

The additive noise is defined in terms of the gradient of a cost function, which is to be minimized in the learning phase. Since the cost function is enriched by total variation and bounded range regularization terms, the generated adversarial images ``look" natural. 

The images generated by the JND method are forced to be consistent with the human visual system, yet they can successfully trick the classifiers on ImageNet and CIFAR10 datasets and the object detectors on MS COCO dataset.

We compare the JND method with three state-of-the-art methods, namely,  FGSM, FGV, and DeepFool, in terms of closeness to the original images and model fooling speed while maintaining the quality of the generated images for image classification and object detection tasks. For the object detection task, unlike FGSM and FGV, the JND method can generate both targeted and non-targeted adversarial examples, which gives the researchers the flexibility of studying both attack methods. 

We conduct experiments on Imagenet and CIFAR10 datasets to analyze the quality of the adversarial images using popular image quality metrics. The results show that the images generated by the JND method are better than the above-mentioned methods.

We conduct experiments on the CIFAR10 dataset to analyze the statistical properties of the adversarial images. The results show that the images generated by the JND method have smaller KL-divergence and $L_2$ distances between the adversarial and original images.

 We also conduct experiments on MS COCO dataset to compare both targeted and non-targeted attack performances on object detectors. We find that all methods are capable of generating non-targeted attacks. However, only the JND method can generate targeted attacks, which shows its superiority over other methods.

Finally, we compare the model fooling speeds while maintaining the quality of the generated images.  We observe that the suggested JND method converges faster than the other methods.

In summary, the generated images with JND look more natural and authentic while tricking the classifiers and object detectors at a faster speed. It outperforms the FGSM, FGV, and DeepFool methods on whole dataset attacks.


\bibliographystyle{unsrt}       
\bibliography{main}   


\end{document}